\documentclass[11pt,a4paper]{article}
\usepackage{emnlp2018}
\usepackage{times}
\usepackage{latexsym}
\usepackage{enumitem}

\usepackage{url}
\usepackage{booktabs}

\aclfinalcopy
\usepackage{graphicx}
\title{Survey of Text-based Epidemic Intelligence: A Computational Linguistic Perspective}
\author{\begin{tabular}{ccc}
Aditya Joshi$^{1}$& Sarvnaz Karimi$^{1}$ &
\end{tabular}\\
\begin{tabular}{ccc}
\textbf{Ross Sparks$^{1}$} & \textbf{C\'{e}cile Paris$^{1}$} & \textbf{C Raina MacIntyre$^{2}$}\\
\end{tabular}\\
\begin{tabular}{ccc}
\multicolumn{3}{c}{$^{1}$CSIRO Data61, Sydney, Australia}\\
\multicolumn{3}{c}{$^{2}$Kirby Institute, The University of New South Wales, Sydney, Australia}\\
\multicolumn{3}{c}{\tt \{aditya.joshi,sarvnaz.karimi\}@csiro.au}\\
\multicolumn{3}{c}{\tt \{ross.sparks,cecile.paris\}@csiro.au, \tt r.macintyre@unsw.edu.au}
\end{tabular}
}

\begin{document}
\maketitle
\begin{abstract}
Epidemic intelligence deals with the detection of disease outbreaks using formal (such as hospital records) and informal sources (such as user-generated text on the web) of information. In this survey, we discuss approaches for epidemic intelligence that use textual datasets, referring to it as `text-based epidemic intelligence'. We view past work in terms of two broad categories: health mention classification (selecting relevant text from a large volume) and health event detection (predicting epidemic events from a collection of relevant text). The focus of our discussion is the underlying computational linguistic techniques in the two categories. The survey also provides details of the state-of-the-art in annotation techniques, resources and evaluation strategies for epidemic intelligence. 
\end{abstract}
\textit{This paper is under review at ACM Computing Surveys. This version of the paper does not use the ACM Computing Surveys stylesheet. We would appreciate your feedback on the paper. Please write to the primary author if you have any suggestions. All figures have been removed because there were formatting issues with the new stylesheet.}
\section{Introduction}
Epidemics have adversely impacted lives and well-being of individuals, and as a result, economies, for centuries\footnote{\url{http://edition.cnn.com/interactive/2014/10/health/epidemics-through-history/}}. While limitations of medical knowledge accounted for most of the impact, delayed detection and latency of prevalent communication channels were a bottleneck. For instance, about a century ago, slow word-of-mouth communication between medical professionals would aggravate the impact of disease outbreaks due to a late detection. The rise of information technology provided a new channel to medical professionals to exchange information and possibly detect anomalous events in the health of a community. Referred to as `\textbf{epidemic intelligence}\footnote{\url{https://www.who.int/csr/alertresponse/epidemicintelligence/en/}}', such systems aim to provide early warnings of public health emergencies.  Using information from digital sources, such surveillance can potentially mobilise a rapid response resulting in reduced rate of morbidity and mortality~\cite{yan2017utility}. Epidemic intelligence has proven to be useful during several instances of outbreaks in the past, including the early detection of an A (H1N1) pandemic~\cite{brownstein2009digital}. An overlapping area of research is syndromic surveillance~\cite{henning2004syndromic,yan2006review}, where the goal is to detect syndromes: a collection of related symptoms. A syndrome is a condition characterised by associated symptoms, while a symptom is an indication of a medical condition\footnote{Source: Oxford Dictionary.}. However, `syndromic surveillance' is no longer restricted to syndromes in its true definition (a collection of related symptoms)~\cite{hopkins2017practitioner}. 

Traditionally, epidemic intelligence relies on structured information from medical institutions and governing bodies such as medical records or weather information respectively~\cite{yan2006review}. The use of internet to obtain information from such official sources has recently been a popular paradigm for epidemic intelligence ~\cite{brownstein2009digital}. Epidemic intelligence has been impacted by the rise of textual content on the internet. In particular, Web 2.0\footnote{\url{https://en.wikipedia.org/wiki/Web\_2.0}} enabled users to generate textual content on the internet. As a result, many informal sources of data (like online discussion forums) are often first reports of an epidemic\footnote{\url{https://www.who.int/csr/alertresponse/epidemicintelligence/en/}}. The data under consideration for epidemic intelligence may span a spectrum of sources: short phrases in the form of queries entered in search engines, or long documents such as news articles or blogs written by users. Intermediate between the two are posts on digital social media (referred to as `social media' in the rest of the paper). Due to its accessibility and popularity, social media is an attractive source of data for epidemic intelligence, like for other fields of analytics. The value of textual data for epidemic intelligence can be understood by the observation that more than 60\% of initial reports of epidemics are received from unofficial informal sources, including text-based sources~\cite{wagner2011handbook}. We refer to \textit{the detection of epidemics using health-related textual data} as text-based epidemic intelligence.

Since it involves textual data, text-based epidemic intelligence uses techniques in computational linguistics/natural language processing (NLP). Text-based epidemic intelligence can be viewed as a two-step process, indicated in Figure X, as follows (Figure will be made available in the full paper):
\begin{itemize}
\item \textit{Health mention classification}: This step refers to the identification of text that is relevant to epidemic intelligence. For example, \citet{aramaki2011twitter} predict if a tweet reports an influenza outbreak. As shown in the figure, this step selects text concerning public health risks of interest, from a large pool of textual data (such as Twitter streams).
\item \textit{Health event detection}: In this step, the extracted information from the relevant text is applied in order to identify health events. An example is the work by~\citet{sparks2010exponentially}, where they use exponentially weighted moving average to predict influenza counts over time.  As shown in the figure, this step aims to detect epidemics by taking into account a collection of textual units (as opposed to one textual unit in the previous step).
\end{itemize}

Figure X (Figure will be made available in the full paper) presents an overview of the research in text-based epidemic intelligence in terms of these steps. Events in the real world get manifested in the form of online textual content such as news articles and social media text. In addition, structured textual data such as medical ontologies provide knowledge about the domain. Health mention classification involves computational linguistic techniques based on ontologies, statistical classifiers or topic models. In contrast, health event detection involves identifying a health event corresponding to a possible outbreak in the community. Temporal outbreaks refer to outbreaks over time where the textual units are arranged in a time series. Spatial outbreaks refer to outbreaks over space where the textual units are arranged in a geographical region. 

Because it aims to select relevant text from large volumes of text, health mention classification has witnessed more diversity in terms of computational linguistic approaches in comparison with health event detection. Hence, we survey different approaches reported for health mention classification, and highlight strategies peculiar to textual datasets that have been used in health event detection.

The survey paper is organised as follows. Section~\ref{sec:motivation} positions this survey paper among related survey papers. In Section~\ref{sec:scope}, we look at different ways in which past works define their scope. Following this, we describe resources for epidemic intelligence in Section~\ref{sec:resources}. We then survey past work in health mention classification in Section~\ref{sec:syndmodeling} followed by health event detection in Section~\ref{sec:syndmonitoring}. Section~\ref{sec:eval} describes the evaluation techniques that have been used. A list of possible future directions is in Section~\ref{sec:openres}, while the conclusion is in Section~\ref{sec:concl}.

\section{Motivation}
\label{sec:motivation}
The earliest literature review of syndromic surveillance using open-source data was by \citet{yan2006review} in 2006. It surveyed approaches of the time, primarily based on structured information sources (such as hospital records). Several years later, other papers applied systematic review techniques to summarise syndromic surveillance~\cite{charles2015using,al2016using}. \citet{bernardo2013scoping} used a structured scoping review method using a dataset of 101 scientific articles, reports statistics in terms of demographic attributes of authors and themes of these publications. A comprehensive list of existing surveillance systems can be found in \citet{yan2017utility}.  Surveys of other specific problems (namely adversely drug reaction detection) in health informatics are by \citet{karimi2015text} and \citet{sarker2015utilizing}. A systematic review closely related to ours is by \citet{velasco2014social}. They provided a high-level view of approaches for syndromic surveillance using social media text. Our computational linguistic perspective appears in terms of viewing past work in epidemic intelligence via typical computational linguistic techniques. Our survey is targeted towards enabling computational linguistic as well as epidemic intelligence researchers to understand the state-of-the-art. The novelty of this survey is as follows:
\begin{enumerate}
\item Our survey paper views epidemic intelligence as an application of computational linguistics and presents approaches for syndromic surveillance that process textual data.
\item Our view of epidemic intelligence divides past work into two steps: Health mention classification and health event detection.
\item We classify past approaches in health mention classification in terms of the class of well-known NLP paradigms. This helps to understand trends in the types of approaches that have been reported over time.
\end{enumerate}

The need for application of advanced computational linguistic techniques arises from typical challenges involved in text-based epidemic intelligence:
\begin{enumerate}
\item \textbf{Term presence is not sufficient}: Ambiguity is a key challenge for natural language processing~\cite{manning1999foundations}, and this holds for epidemic intelligence as well. A sentence containing a symptom/illness may not always be the report of the illness. For example, `\textit{I have the flu}' is a report of an illness, while `\textit{Flu is common in winters}' is not. Therefore, health mention classification must be able to distinguish between health reports (where a person reports experiencing certain symptoms) and other tweets. This maps itself to a two-class classification task. The classification becomes more challenging if it is more fine-grained. For example, it may be useful to classify if a given sentence is a suspicion (`\textit{I have a head ache. Maybe I have a flu}'), a fact (`\textit{WHO reported that bird flu is likely this year}') or a question (`\textit{Do you have flu?}').
\item \textbf{Targets may be important}: The target of a health mention is the person who has contracted the disease. For example, in `\textit{my flu got cured}', the target is the speaker, while in `\textit{my mother has been down with a flu}', the target is the mother. In case of health event detection, target may play an important role. This means that, in addition to detecting the presence of a health mention, it may be required to detect who has the health mention. If the user mentions their flu in many tweets or if a famous person falls ill, there may be a high number of health mention tweets but it may not always warrant signalling a health event. \citet{kanouchi2015caught} present an approach to detect the target of a tweet as one among 1st person, 3rd person, referred person (in the tweet @target, for example), not human, and none. They deal with seven symptoms: cough, cold, headache, chill, runny nose, fever, and sore throat. The target of a health mention may be challenging to determine due to typical challenges in social media text: (a) The subject may be somebody else (`\textit{my brother got flu}'), (b) The verb may indicate the target (`\textit{my brother has passed his flu to me}'), or (c) subject may be dropped (`\textit{Awake with a headache}').
\end{enumerate}
Given these challenges, this survey presents nuances of computational linguistics techniques in terms of annotation strategies, approaches and evaluation techniques for epidemic intelligence.

\section{Scope Definition}
\label{sec:scope}
We now describe how past papers define their scope. For example, some papers focus on tweets about a mass gathering where there may be a risk of an outbreak, while some other focus on specific contagious diseases. In this section, we discuss several such dimensions that could be useful to formulate a research problem in text-based epidemic intelligence.
\citet{gomide2011dengue} describe four requirements of a epidemic intelligence system: how much (extent), where (location), when (time) and how (manner of spread). To this list, we add `what' to refer to the health condition of interest. We discuss the scope of past work in light of these requirements.

\subsection{Illness (`\textit{What}')}
The `what' of an epidemic intelligence system is the health condition of interest. This could be an illness characterised by symptoms or a syndrome. Instead of a generic, illness-agnostic system, past work focuses on specific symptoms. A scoping review shows that most past work deals with influenza and influenza-like illnesses~\cite{bernardo2013scoping}. Other forms of illnesses that have been studied include sexual health, alcoholism, and drug abuse~\cite{charles2015using}. In general, the parameters that influence the choice of syndrome in a study are:
\begin{enumerate}
    \item \citet{karisani2018did} state that symptoms that are apparent to a patient can be a good choice if social media datasets are to be used. They refer to text reporting such symptoms as a `\textit{personal health mention}'. The goal of such research is to identify if a given piece of text contains a person reporting an illness or not.
    \item Since the epidemic intelligence approaches need to be validated, the availability of reference counts from official records are important determiners. For example, counts from the Center for Disease Control and Prevention (CDC) or its equivalent in other countries have been used as a source of validation sets~\cite{boyle2011prediction}.
    \item Social stigma and privacy concerns arising due to it may prevent certain illnesses from being discussed on social media~\cite{fung2015use}. In such cases, the volume of social media content reporting the illness may be insufficient to detect an epidemic.
\end{enumerate}

\subsection{Time period (`\textit{When}')}
The second parameter is the time period. One way to view the time period is in relation to a known outbreak. \citet{sparks2017investigation} provide two categories, using this relativity:
\begin{enumerate}
\item \textbf{Retrospective surveillance}: Retrospective surveillance looks at an outbreak in retrospect in order to understand past unusual behaviour, using validation data of past outbreaks. Insights from retrospective surveillance can have an impact on prospective surveillance.
\item \textbf{Prospective surveillance}: This kind of surveillance involves monitoring for health indicators and triggering appropriate flags, if an outbreak is detected. These systems are time-critical in that they aim to detect outbreaks as early as possible~\cite{sparks2010understanding}. Mobile phones have been used for prospective surveillance to ensure that the detection of health signals is within an acceptable time period~\cite{rosewell2013mobile}.
\end{enumerate}
\subsection{Location (`\textit{Where}')}
Some past work also define in their scope a specific event. These are typically high-risk events that may trigger medical emergencies. This may, as a result, cause people to post on the web to express fears, report symptoms, etc. The objective of such work is then to harness this web content to detect these outbreaks. For example, \citet{brownstein2009digital} discuss how epidemic intelligence was useful during an early A(H1N1) pandemic. Other event-based focuses have been an ebola outbreak in London~\cite{ofoghi2016towards}, a Zika virus outbreak around the world~\cite{adam2017zikahack} or the 2002 Winter Olympic games~\cite{chapman2005classifying}. \citet{henning2004syndromic} refer to this kind of surveillance as short-duration or drop-in surveillance, since it is centered around an interval of time for which the event lasts. In addition, some work also focuses on specific cities~\cite{chapman2005classifying}, or trains systems on multiple geographical locations~\cite{zou2018multi}.
 
\subsection{Indirect Indicators (`\textit{How much}')}
Some research estimates the extent of impact of an outbreak using indirect indicators. \citet{ofoghi2016towards} relate public health threats to public mood about disease names. The hypothesis is that real-world health threats may be discovered by detecting emotions in related social media text. Thus, the context in focus in this case is the sentiment about a syndrome as against the actual incidence of the syndrome. In order to detect the public mood, they formulate an emotion analysis task on health-related tweets, and use it to identify possible threats. Similarly, \citet{larsen2015we} use an emotion analysis tool to detect emotions in social media text as an indicator of health signals.

\subsection*{A Note on Ethics}
In addition to the requirements that help to define scope of a epidemic intelligence research, a note on ethical considerations is imperative. \citet{benton2017ethical} state that public datasets are exempt from being regarded as sensitive.They also state that social media may be useful since it is potentially a public dataset. They provide a starting point for researchers for ethical guidelines. \citet{benton2017ethical} also describe the importance of ethical guidelines for health surveillance research using social media. They prescribe mechanisms such as an institutional review board, informed consent from participants, and protection of sensitive data. Similarly, \citet{ginsberg2009detecting} describe techniques like anonymisation of health data using identifiers, or the use of normalised counts instead of specific instances. A detailed discussion on ethics is beyond the scope of this survey. However, papers like 

\section{Resources}
\label{sec:resources}
 Textual resources are the foundation of a computational linguistic system~\cite{joshi2017sentiment}. Structured resources include ontologies or lexicons that provide semantic information about the domain. Unstructured resources include labeled datasets that contain instances assigned with labels of interest.  
\subsection{Ontologies}
\label{sec:ontologies}
 An ontology is a formal, explicit specification of a shared conceptualisation~\cite{ontologydef}. Ontologies consist of concepts, and relations that link two or more concepts. Medical ontologies have helped to organise the knowledge of the medical domain~\cite{bertaud2012ontology}. It must be noted that, apart from epidemic intelligence, these medical ontologies have been used for applications based on medical information extraction or information retrieval.

The Unified Medical Language System (UMLS) provides a popular medical ontology~\citep{lindberg1993unified}. This is a meta-thesaurus that defines biomedical concepts and relations between them. UMLS was created by experts using information captured in multiple related ontologies such as a gene ontology. UMLS captures relationships of two types: associative (such as \textit{has}) and hierarchical (such as \textit{isa}). Associative relationships may be used to relate symptoms with a syndrome where the symptoms are observed. Hierarchical relationships may represent a specialisation chain of illnesses.  Other biomedical sources have also been associated with concepts and relations in UMLS. \citet{bodenreider2004unified} describe tools that allow customisation and enrichment of the UMLS ontology. 

\citet{collier2007biocaster} describe an ontology that captures syndromic knowledge. This ontology, known as the BioCaster ontology, consists of: (a) concepts such as disease, symptom, virus, or syndrome, and (b) relations such as $has\_symptom$ that relates a disease with a symptom, or $causes$ that relates a disease with the virus that causes the disease. The ontology is multilingual with support in 12 languages, including but not limited to English, Japanese, French, Arabic and Thai. 

\citet{okhmatovskaia2009sso} present a syndromic surveillance ontology (SSO) for certain classes of illnesses: respiratory, gastrointestinal, constitutional and influenza-like. The ontology allows capturing definitions of a higher granularity for syndromes. \citet{conway2011developing} report an extended version of the SSO that covers a broader range of illnesses.
\begin{table*}[tb]
\begin{tabular}{p{2cm}|p{2cm}|p{4cm}|p{4cm}}
\toprule
\textbf{Data Source}       & \textbf{Nature of Text} & \textbf{Advantages}    & \textbf{Challenges}   \\\midrule
Search queries~\cite{zou2018multi,ginsberg2009detecting,hulth2009web} & Short phrases by users of search engines & Search queries can be aggregated in terms of counts, allowing large-scale monitoring & There may not be a direct correlation between a need for information and occurrence of an outbreak. Also, search queries may not be readily available. \\ \midrule
News articles~\cite{doan2008global,yangarber2008content,lejeune2010filtering,freifeld2008healthmap}    & Well-formed text written by journalists & High-quality NLP tools such as parsers, part-of-speech (POS) taggers can be used.                     & News articles may contain latency because they may be written periodically. \\\midrule
Medical reports~\cite{crubezy2005ontology,conway2011developing,aamer2016syndromic,olszewski2003bayesian}   & Text written by medical experts   & They represent reliable information reported by medical professionals.               & There may be privacy concerns to obtain these datasets. The short nature of the text may make it difficult to infer information.                                                 \\\midrule
Social media text~\cite{yepes2015investigating,adam2017zikahack,lampos2017enhancing,ofoghi2016towards} & Short text written by users of social media      & The frequency and volume makes it an attractive source of data. & The text may be noisy and unreliable. This may result in false signals.\\
 \bottomrule
\end{tabular}
\caption{\label{tab:datasources}Summary of Unstructured Data Sources.}
\end{table*}

While medical ontologies use different representations and may encompass illnesses, they play a key role as a knowledge base for several epidemic intelligence approaches. It must be noted that medical ontologies capture medical names as well as colloquial names of symptoms. This becomes crucial because health mentions in informal text such as social media may not contain scientific/technical terms.
\subsection{Datasets}
\label{sec:datasets}
In the previous section, we described ontologies that provide a structured background knowledge for epidemic intelligence. In this section, we present labeled datasets used for epidemic intelligence. We first describe the sources from where the data is obtained, and annotation strategies that have been employed to obtain annotations.
\subsection{Data Sources}
 By data sources, we refer to different classes of textual content that may be used to create data for epidemic intelligence. Each of these data sources offers interesting opportunities and poses specific challenges. \citet{velardi2014twitter} categorise these sources as follows:
\begin{enumerate}
\item \textbf{Demand-based data sources}: This refers to sources that reflect demand for information. A popular type of demand-based data source is search engines. In this case, the assumption would be that a large volume of search queries is likely to indicate a prevalent health risk in the community. However, demand-based data sources may not provide good estimates of health risks of interest. For example, in the case of search queries, not all searches may not be linked to a personal symptoms. In the wake of the outbreak of a disease, media coverage may result in fears in the minds of people resulting in a higher demand for information regarding the disease~\cite{alicino2015assessing}. In addition, demand-based data sources may not be readily available.
\item \textbf{Supply-based data sources}: Supply-based sources are the ones where the data originates on large-scale platforms designed to share information. Examples of such platforms are discussion forums and social media. While such platforms may provide large-scale information, the text tends to be longer than search queries. Extraction of relevant textual items from a large pool is a key challenge with supply-based data sources. This has motivated the majority of the research in health mention classification, where labeled datasets from supply-based sources are used to learn systems for this classification. 
\end{enumerate}
Datasets originate from these two categories of sources, as shown in Table~\ref{tab:datasources}. They are:
\begin{enumerate}
\item \textbf{\textit{Search Queries}}: Search queries provide a large-scale view of interests of people express, along with reasonable anonymity \cite{hulth2009web}. Therefore, search queries are an attractive data source for text-based epidemic intelligence. A seminal work by \citet{ginsberg2009detecting} describes a system which uses volumes of queries on Google, to detect disease outbreaks. Popular as Google Flu Trends, this system uses counts of search queries to predict influenza-like infections (ILI). However, in recent times, Google Flu Trends has received criticism for over-estimating flu counts due to modifications in Google's search algorithms\footnote{\url{http://time.com/23782/google-flu-trends-big-data-problems/}}. 
    \item \textbf{\textit{News Articles}}: News feed monitors based on RSS feeds (\url{https://en.wikipedia.org/wiki/RSS}) enable access to news websites in various languages. This allows monitoring them for news articles. Since news articles are typically written by professional writers, they adopt a formal style of writing. This makes it possible to use NLP tools such as semantic taggers and parsers to extract information about the health incidents. Early work in epidemic intelligence relies on news articles. \citet{doan2008global} present Global Health Monitor, a system that uses news feeds to identify health issues. This system uses a pipeline of NLP tools to identify health risks that communities may be facing. The system periodically scans 1500 news feeds for relevant content. \citet{yangarber2008content} also rely on news articles to detect health outbreaks. In each of these cases, a set of pre-determined keywords corresponding to a set of illnesses are used to monitor the news feeds. In addition to using news articles as a source of data, peculiarities of news articles can also be leveraged. \citet{lejeune2010filtering} show how news articles from multilingual sources can be monitored for health mentions. Their approach relies on a popular journalistic style where the head of a news article contains details of a health incident in terms of location, time, etc. Using a set of rules based on entities in the head and the body of a news article, they fill a template that corresponds to information about a health incident. Their experiments are conducted on news articles in English, Chinese and French. \citet{freifeld2008healthmap} describe Health Map, a system that uses news reports to monitor diseases.
    \item \textbf{\textit{Medical reports}}: Reports from medical sources have also been used as textual datasets. Chief complaints are primary reports created by emergency departments of hospitals~\cite{conway2013using}. They are often short strings that describe the medical condition of the patient when they first report to a hospital. On similar lines, \citet{crubezy2005ontology} use 2256 records in a medical record repository. \citet{olszewski2003bayesian} use a dataset of 28,990 triage diagnoses (ranging from 1 to 10 words in length). \citet{aamer2016syndromic} use the SynSurv dataset that contains 2006 reports from two Melbourne hospitals.

\item \textbf{\textit{Social Media Text}}: Online social media such as Twitter allow users to post text. Availability of APIs to access Twitter have been useful. \citet{yepes2015investigating} state that social media provides targeted health information without the legal and technical obstacles that exist for other sources such as official records. However, \citet{adam2017zikahack} point out that social media text may not be as credible or reliable as official records. \citet{lampos2017enhancing} use a dataset of 35000 tweets posted over 449 weeks. \citet{ofoghi2016towards} collect three datasets of tweets. The first two are related to an outbreak of Ebola: pre-event (\textit{i.e.}, tweets before an outbreak) and post-event (\textit{i.e.}, tweets after an outbreak). In addition to the two, they also use an Ebola background dataset. This dataset is about Ebola but is long before any outbreak occurred. Therefore the third dataset acts as a negative dataset which, although contains mentions of Ebola, is not related to a health outbreak. They obtain tweet-level manual annotations. Each tweet is labeled with emotion classes such as happiness, criticism, disgust and sarcasm.
\item \textbf{\textit{Combination}}: Some past work combines textual datasets from different data sources. This is either to validate that an approach holds for different text forms or for the information from multiple sources to supplement each other. \citet{neveol2009exploring} experiment with two datasets: 551 sentences from medical literature, and around 500 queries from the PubMed website. \citet{woo2016estimating} use data from multiple textual sources in Korean: search queries, social media data such as blogs, and correlate it with national influenza data.
\end{enumerate}

\subsection{Dataset Considerations}
 To create labeled datasets for health mention classification, datasets must be annotated with labels of interest. In such cases, one must consider: (a) what are the instructions to the annotators?; (b) how is the quality of annotations ensured/validated? It must be noted that the strategies described below are implemented in conjunction with each other, and not in isolation.

In terms of obtaining annotated datasets, appropriate guidelines to annotators is key. \citet{aramaki2011twitter} create a dataset of 5000 training tweets, manually labeled as positive/negative for the task of detecting health mentions. The annotator guidelines state that a tweet should be labeled positive if: (a) one or more people with flu exist around the tweet author; and (b) the tense is present or recent past. The authors also mandate that the tweet should be affirmative and not speculative (for example, `\textit{Seems like I might have flu}').

Despite the annotation guidelines, some annotators may not perform well due to various factors. To identify reliable annotators, \citet{lamb2013separating} create a gold dataset of tweets whose labels are known. Then, manual annotations are then obtained for around 12000 tweets from multiple annotators. These tweets include the gold dataset. Annotations by annotators with greater than 60\% accuracy on a gold dataset are retained. This annotation strategy is illustrated in Figure X (Figure will be made available in the full paper).

Because health event detection deals with a large volume of data generated over a period of time, it may not be possible to obtain annotation for all instances. Therefore, a combination of manual and automatic annotation has also been used.  Figure X illustrates a typical combination. (Figure will be made available in the full paper) In general, the classifier is trained on a small set of manually annotated instances, and predictions are obtained on the complete dataset. These predictions are used as annotations for the dataset. \citet{paul2011you} download a set of 2 million tweets. A subset of 5,128 tweets are manually labeled. A classifier is trained on these manually labeled tweets and the predictions on the remaining tweets are obtained. These predictions are then used as labels for the tweets. A similar technique is used to obtain annotations for a large dataset of around 11 million tweets by \citet{paul2012model}.

\citet{sadilek2012predicting} use a more sophisticated approach of obtaining a combination of manual and automatic labels. Since geolocations are crucial to their approach, they first identify 6237 users who have turned on geotagging. The annotation is then carried out as follows. As the first step, a set of tweets are manually labeled. Then, two classifiers are trained. The first classifier is trained with a high mis-classification cost for  the majority class. This implies that it is expected to do well on the majority class. On the contrary, the second classifier has a high mis-classification cost for the minority class. Both the classifiers are used to obtain predictions on the unlabeled portion of the dataset. Predictions with high confidence by either of the two classifiers are added to the labeled dataset. Another method of combining automatic and manual annotation is used by \citet{jiang2016construction} to create a dataset of personal health mentions. They employ an iterative algorithm that begins with a seed set of manually labeled instances. A classifier is trained on these labeled instances and predictions are obtained for an unlabeled set of instances. The labeled instances are then randomly selected for manual annotation. Following manual verification of the labels, these samples are added back to the seed set. The process repeats until the data imbalance is within an acceptable threshold. 

\begin{table*}[t]
\small
\begin{tabular}{p{2cm}|p{3.5cm}|p{3.4cm}|p{3.4cm}}
\toprule
\textbf{Approach} & \textbf{General Idea} & \textbf{Motivation}   & \textbf{Challenges}    \\ \midrule
Ontology-enhanced~\cite{collier2010ontology, huang2016syndromic,lu2009multilingual,crubezy2005ontology, conway2013using} & Given a text, map the terms in the text to appropriate concepts in the ontology to determine if a syndrome can be detected. & Medical ontologies capture useful information in a structured form. & (A) Ontologies may not be complete, (B) Ontologies may contain medical terms while the text may contain colloquial terms. \\\midrule
Similarity-based~\cite{freifeld2008healthmap, aamer2016syndromic, ofoghi2016towards} & Similarity between distributions and similarity between concepts are used as indicators of an illness.  & A text that is similar to illness concepts/text is likely to be about the illness. & The choice of similarity metric determines the benefit. \\\midrule
Topic Model-based~\cite{wang2014exploring,chen2016syndromic, paul2012model, paul2011you}  & With the help of datasets from social media topic models that are extensions of Latent Dirichlet Allocation (LDA) model have been proposed. With the use of additional latent variables, these models provide structured information about illnesses.& Topic models can process unlabeled/partially labeled data and provide valuable information. & Interpretation of generated topics and their application to Health mention classification may not be straightforward.  \\\midrule
Pipeline-based~\cite{doan2008global,yangarber2008content,yepes2015investigating,yates2014framework}   &  These approaches combine existing NLP components to build effective deployments. Typical components include named entity extraction and text classification. &  Health mention classification can be broken down into a sequence of NLP components fitting into one another. &  NLP components may be trained on documents in domains unrelated to health-care. In such cases, their efficacy for health mention classification needs validation.   \\\midrule
Statistical~\cite{olszewski2003bayesian,aramaki2011twitter,chapman2005classifying,lamb2013separating,kanouchi2015caught,jiang2016construction}   & Features based on words, emotion scores, medical concepts and POS tags, along with typical classifier learning algorithms have been reported.  & Supervised classifiers trained on labeled datasets have been found to be useful in many applications of NLP.   & Selecting appropriate features and ensuring they generalise may be challenging.   \\\midrule
Deep Learning-based~\cite{karisani2018did,dai2017social,lampos2017enhancing,wang2017using}  & Features based on word embeddings and modification of general-purpose word embeddings to the specific domain space have been reported, along with typical neural network models. & Deep learning approaches have proven to be useful since they do not rely on human-engineered features. & Lack of availability of large labeled datasets may be an impediment.  \\  \bottomrule
\end{tabular}
\caption{Summary of Approaches for Health Mention Classification.}
\label{tab:summarymodeling}
\end{table*}
\section{Health Mention Classification}
\label{sec:syndmodeling}
In the survey so far, we introduced the problem of epidemic intelligence, motivated it in terms of its challenges and then described the resources that can be used for epidemic intelligence. This section describes text-based approaches that have been reported for health mention classification. We classify past approaches in categories that are commonly used in computational linguistics, summarised in Table~\ref{tab:summarymodeling}. We detail out these approaches in the forthcoming subsections.
\subsection{Ontology-enhanced Approaches}
In ontology-enhanced approaches, an ontology provides relevant medical knowledge that is used to make appropriate predictions. In general, an ontology can play the following roles in a epidemic intelligence system:
\begin{itemize}
\item To extract entities of different types using patterns (for example, `\textit{X leads to Y}' can be used as a pattern to infer that X is a cause of an illness Y);
\item To identify diseases based on their common name, medical name or symptoms; and
\item To use inference rules from medical-domain ontologies to predict medical events of interest
\end{itemize}

An application that uses the BioCaster ontology  is reported by
\citet{collier2010ontology}. Their system monitors news feeds for health-related news. They monitor news articles from multilingual sources. Based on the target keywords, relevant news articles are then translated to English. Following this, they use topic classification to further filter news relevant to the medical domain. For these, they then use information extraction techniques, such as named entity recognition or semantic role labeling, to construct relationships in the BioCaster ontology. Based on the relation tuples derived from the ontology, the system predicts public health events.
\citet{crubezy2005ontology} combine concepts in two ontologies by measuring relatedness between them. With this concept mapping, they classify a chief complaint into one of many syndrome categories by using a rule-based inference technique. A problem solver carries out the inference over the ontologies. They report that 44\% of errors that were analysed are due to concept mapping not being found.  \citet{lu2009multilingual} use an ontology along with cross-lingual projections for classification of chief complaints in Chinese. A chief complaint is first pre-processed to account for stylistic properties of Chinese script and language. The words are projected to English using translation, and a chief complaint classifier trained on English documents is used for classification. Then, significant terms related to the medical domain in the complaint are identified and matched with those in an ontology. \citet{conway2013using} perform a review of chief complaint classification systems in North America. These systems use a combination of ontology-enhanced and statistical approaches. \citet{huang2016syndromic} use a medical ontology which contains associative relationships between medical concepts, \textit{i.e.}, information on how these concepts related to each other. To use this ontology, if a word in a tweet is predicted as an entity of interest, it is mapped to a concept present the ontology using similarity values. The concepts themselves become the features of a classifier that detects an illness.
\subsection{Similarity-based Approaches}
Similarity-based approaches use notions of similarity to model syndromes. In general, the idea is to obtain semantic distances between words in a text and terms related to a syndrome of interest. Several similarity-based approaches have been reported. \citet{freifeld2008healthmap} use an N-gram-based approach that matches n-grams in a news report with those in a known dictionary of terms, based on semantic distances. Based on this matching, they classify every news report in terms of two parameters: primary location and disease name. \citet{aamer2016syndromic} present a semi-supervised algorithm that uses similarity to an illness-related concept as an illness-related indicator. They use Jenson-Shannon divergence to compute the similarity between terms for a dataset of chief complaints. In order to filter terms, they use a log-likelihood-based technique. 
Similarly, \citet{ofoghi2016towards} use Na\"{i}ve Bayes and lexicon-based approaches. They report Kullback Leibler (KL) divergence between emotion class distributions for pre-event and post-event datasets.
\subsection{Topic Model-based Approaches}
Topic models allow discovery of thematic concepts underlying large datasets. Latent Dirichlet Allocation (LDA) model is a popular topic model based on the assumption that a document is composed of a mixture of concepts (referred to as `topics')~\cite{bleispaper}. While LDA models have been used to obtain themes underlying health-related datasets, there have been two extensions of the LDA model designed to understand aspects of syndromes. The first topic model is by \citet{paul2012model}, called the Aspect Topic Ailment Model (ATAM). The model includes a latent label each for: (a) switching between general or health-related words; (b) identifying background words; and (c) an ailment. Using an observed label to select between ailment, treatment and general health-related words, the model discovers topics corresponding to ailments. A stochastic version of the Expectation-Maximisation algorithm is used for the estimation of latent variables in the model, to maximise the likelihood of the data. While this work focuses on flu, an extension of this work by \citet{paul2011you} reports findings on a wider range of symptoms. \citet{wang2014exploring} use the ATAM to extract topics from Chinese micro-blogs, and discover topics corresponding to health. The second extension of a LDA model is by \citet{chen2016syndromic}. This model is called the Hidden Flu-State Tweet Model (HFSTM). It uses the Twitter timeline of a user as a temporal series. For each tweet, the model estimates the health state of a user as: healthy, exposed and infected. It uses latent variables similar to the ATAM: (a) a word-level variable that indicates background words, (b) a word-level variable that indicates general domain words, (c) a word-level switch variable between symptom and general words, and (d) a tweet-level symptom variable. The symptom variable for a tweet, in addition to the local word-level dependencies, depends on the symptom variable of the previous tweet in a sequence. This way, the model incorporates temporal property in Twitter timelines. In the case of all approaches, the datasets are created using symptom keywords, so as to ensure that the topics are relevant. The following holds for both these topic model-based approaches:
\begin{itemize}
\item Additional latent variables and dependencies are constructed to incorporate semantic relationships between types of word clusters. These relationships may be in the form of symptoms, ailments and medication for a syndrome, where symptoms, ailments and medication have topics corresponding to each.
\item If a list of words indicating infections and another indicating medicines are available, asymmetric priors may be set on these words for a certain set of topics. This appears to be helpful to discover other unrelated symptoms, if a smaller set of symptoms is known based on medical expertise.
\end{itemize}

\subsection{Pipeline-based Approaches}
The next category of approaches is the pipeline-based approaches. We call them such because these approaches present solutions in the form of a pipeline of computational linguistic modules. Some pipeline-based approaches for epidemic intelligence are as follows. \citet{doan2008global} use a three-step approach in their news monitoring system: (a) Topic classification using a Na\"{i}ve Bayes classifier first identifies if a tweet is health-related, (b) Named entity recognition is used to extract entities. (c) Disease and location detection to extract these terms, (d) Visualisation to represent the news on a map.
 \citet{yangarber2008content} describe a media monitor for healthcare called MedISys. The system is as an information retrieval engine for medical domain, as a part of the Europe Media Monitor. The system operates as follows: (i) It first searches news articles from feeds. The news articles are then categorised as health-related or not. (ii) For the news articles that are predicted as health-related, they run an information extraction system called PULS. PULS uses a pattern-based technique to extract incidents. (iii) MedISys eliminates redundancies because of the same incident being reported at multiple places.	An incident is defined by four sets of attributes: (a) Location, (b) Disease name, (c) Date/Period, (d) Victim information. Victim information is characterised by features such as Type (human, animal), number, whether survived or not. \citet{yepes2015investigating} describe a system that identifies tweets related to illnesses of interest, and then places the tweet on a map. The pipeline consists of the following modules: (i) Medical Named Entity Recognition (NER) tagger (using a Conditional Random field (CRF), labels words as one of disease, symptoms and pharmacological substances), (ii) Geotagger (If a Global Positioning System (GPS) location is not present, it uses a gazetted list and the user profile location to tag tweets with location). \citet{yates2014framework} describe a framework for epidemic intelligence using social media. Their framework also consists of a pipeline of three steps: concept extraction, concept aggregation and trend detection. In the concept extraction step, they use taggers for named entity recognition and information extraction. In concept aggregation, they identify relationships between the concepts extracted in the previous step. To do so, they use an approach based on word sense disambiguation where different concept words map to the same concept. The third step is trend detection where they plot these counts on a timeline. In general, a typical pipeline-based approach consists of:
\begin{itemize}
\item Use an information extraction tool to get terms of interest.
\item Train machine learning models for relevant predictions.
\item Map these to appropriate data structures.
\end{itemize}

\subsection{Statistical Approaches}
Statistical approaches model health mention classification as a supervised classification problem. In order to describe these approaches, we consider two parameters: (a) the learning algorithm, and (b) the features used to represent an instance\footnote{In this case, we consider as statistical, the approaches that require feature engineering. Deep learning-based approaches are covered in the next subsection.}. 

\citet{olszewski2003bayesian} use a Na\"{i}ve Bayes classifier for prediction of class of illnesses. They use unigrams and bigrams as features of the classifier. \citet{chapman2005classifying} use a probabilistic Bayesian parser to generate semantic frames from chief complaints. These semantic frames are then used to predict presence of a set of diseases. \citet{aramaki2011twitter} use features based on bag of words with feature windows of multiple sizes. They report results on a variety of classifiers such as AdaBoost, Na\"{i}ve Bayes, and support vector machines (SVM). \citet{lamb2013separating} propose features for the task of  infection report detection such as: (a) manually-created set of word classes, (b) tense, person, (c) words indicating concern and awareness, (d) POS n-grams, (e) emoticons, (f) tuples of subject, object, verb combinations. \citet{neveol2009exploring} perform disease detection in medical literature and search queries. They use a technique called the priority model which assigns probabilistic estimates for every query term to belong to either of the two classes: disease mention and disease non-mention. \citet{kanouchi2015caught} use features such as unigrams, weblinks, word classes, length, ngram and retweets for identification of target in tweets that mention health concerns. \citet{jiang2016construction} train a classifier that predicts if a given tweet is a personal health experience. Towards this, they use features such as emotion words, emotion scores, user mentions, and number of first/second/third person pronoun mentions.

\subsection{Deep learning-based Approaches}
The last class of approaches uses neural models based on deep learning. Deep learning allows the discovery of underlying task-relevant semantics without use of human-engineered features. At the heart of deep learning are distributional representations known as `embeddings'. Learned from neural models, a word embedding captures the semantic properties of the word. \citet{lampos2017enhancing} use word embedding distances to select unigram features for flu detection. They use two types of word embeddings: (i) embeddings from wikipedia articles, and (ii) embeddings learned from 215 million tweets from 2014-16.  \citet{dai2017social} use word embeddings to create concept clusters that are then used to classify tweets as flu-related or not. Specifically, they compute disease vectors based on related terms. In order to make a prediction, they create semantic clusters of words using word embeddings such that, for every word, the algorithm randomly chooses between: creating a new cluster, or to adding it to an existing cluster. If any cluster in a tweet is within a threshold of distance to the disease clusters, the tweet is predicted as flu-related. The clustering-based approach is compared against Na\"{i}ve Bayes classifiers. On similar lines, \citet{karisani2018did} present a word embeddings space partitioning and distortion (WESPAD) model for detecting if a given tweet contains a personal health mention. Towards this, they augment word embeddings to other traditional features, and show that the use of word embeddings result in an improvement. However, since the authors believe that word embeddings may not be well-separated for the task, they suggest two innovations: (a) Partitioning: depending on confidence values for each class, they partition the space of embeddings. For each partition, they add two additional features for positive and negative class. (b) Distortion: instead of averaging word vectors to get sentence vectors, they create the sentence vectors by applying info-gain-based weighting to the word vectors.  Finally, \citet{wang2017using} employ an architecture based on recurrent neural networks (RNNs) and compare it with statistical baselines based on classifiers such as SVM.
\subsection{Shared Tasks}
In the context of text-based epidemic intelligence, the following shared tasks have been conducted. \citet{adam2017zikahack} describe a hackathon called ZikaHack held in 2016. The objective was to perform a retrospective analysis of the Zika (also known as the microcephaly virus) outbreak based on different social media sources. The authors describe the systems that participated in the competition. These systems use textual data from sources such as Twitter, Facebook, Instagram, Google maps, Wikipedia, Government reports, in addition to structured data based on medical counts.  The winning system uses translation systems to collect information from multilingual sources, and change point detection algorithms to identify outbreaks.
 \citet{sarker2016social} describe three related tasks: adverse drug reaction detection, drug reaction type classification, and drug consumption normalisation. The competition reported in \citet{weissenbacher2018overview} describes four shared tasks: drug mention detection, medication intake classification, adverse drug reaction detection and vaccination behaviour detection.
\begin{table*}[tb]
\begin{tabular}{p{2.5cm}|p{3.4cm}|p{3.4cm}|p{3.4cm}}
\toprule 
\textbf{Approach} & \textbf{General Idea} & \textbf{Motivation}   & \textbf{Challenges}       \\ \midrule
Temporal Outbreaks \cite{ginsberg2009detecting,sparks2017investigation, hayate2016forecasting,velardi2014twitter, aamer2016syndromic, huang2016syndromic}     & Time series analysis is performed on text bearing timestamps.  & Health events such as epidemic outbreaks can be detected using an unexpected rise of fall in text with certain content.  & A spike may not necessarily correspond to an event. Stigma about certain illnesses may prevent people from writing about it.   \\\midrule
Spatial Outbreaks ~\cite{sadilek2012predicting,chapman2005classifying,ofoghi2016towards,gomide2011dengue,shao2017efficient}    & Location information may be filter relevant text from a region or as features that account for location. &  Location information in text can be helpful to identify possible locations of outbreaks or focus on regions of interest. & (A) Location may not always be available, (B) Outbreaks may often capture interest about the illness around the world, without an actual outbreak in that part of the world. \\  \bottomrule
\end{tabular}
\caption{Summary of Approaches for Health Event Detection.}
\label{tab:summarymonitoring}
\end{table*}
\section{Health Event Detection}
\label{sec:syndmonitoring}
In the previous section, we described approaches that look at the detection of symptoms and syndromes. We referred to them as health mention classification. The second crucial component of epidemic intelligence is health event detection. This refers to approaches that have been used to predict events from a collection of textual units, in terms of time and space. Table~\ref{tab:summarymonitoring} summarises the approaches for health event detection.

\subsection{Temporal Outbreaks}
Detecting a temporal outbreak involves detecting anomalies in the trend of a sequence of textual units. This means that the textual units would need to bear timestamps. Also, the prediction for individual textual units need not be accurate as long as the overall distribution sufficiently points towards an outbreak. There are two broad approaches dealing with temporal outbreak detection: prediction of infection counts (where a number is predicted) and outbreak detection (where the algorithm needs to predict known events).

The first set of approaches predict infection counts from series of textual data. \citet{ginsberg2009detecting} use search counts of the 45 million top queries across a subset of states from the US.	Weekly counts of top queries are normalised by the total number of queries. These counts are stored for every week. Then, they use a model that is trained to predict the influenza-like illness (ILI) counts, given the search query proportions.
 \citet{sparks2017investigation} predict tweet counts using a Poisson regression model that uses features such as hour, day of week, day number in a sequence, as seen in past data. They use process control techniques to detect events depending on whether actual tweet counts are within acceptable range of the predicted/expected tweet counts. \citet{woo2016estimating} use support vector regression to predict influenza counts using numerical features derived from keyword mentions in textual data such as blogs and search queries. \citet{hayate2016forecasting} use a frequency-based approach that factors time lag for different words. For example, they observe that the word `\textit{injection}' lags behind actual outcome of influenza much more than the word `\textit{fever}'. Therefore, they construct the word-day frequency matrix and shift the vector of a word so as to maximise its cross-correlation with the reported flu counts.
 
 The second set of approaches aim to predict known outbreaks. \citet{aamer2016syndromic} perform experiments that consider three formulations: (i) Intersecting seven-day windows, (ii) Disjoint 7-day windows, (iii) Disjoint 1-day windows. They experiment with three degrees of sensitivity - depending on how much the divergence can be. The degrees of sensitivity show the intensity of a likely outbreak. Similarly, \citet{huang2016syndromic} show how the prevalence of flu and Lyme can be detected one week ahead of reported CDC data.	They download English tweets of interest based on keywords from a pre-determined period. Then, they perform named entity recognition to label text with the named entities. These named entities are then labeled for one among disorders, symptoms, and pharmacological substances. The entities in a tweet are mapped to corresponding clusters using an ontology. This allows for medical names and multiple common names to be mapped to the same feature. \citet{velardi2014twitter} use a two-step approach. As the first step, they employ a term extraction algorithm. This algorithm starts with a seed set of technical words and symptom words. This seed set is then iteratively expanded using pattern-matching. This is first done on Google snippets, wikipedia and a medical corpus where related technical words and symptom words are learned. Then, the matching step is repeated on micro-blogs where only symptom words are learned. When the terms are extracted, they count social media mentions of these words as indicators of a syndromic outbreak. 

\subsection{Spatial Outbreaks}
Specifications regarding space have also been considered for health event detection. This may be done either to focus on a certain geographical region, or to use such information as an additional context. To focus on multiple regions, \citet{zou2018multi} use multi-task learning where different geographical regions constitute tasks in a disease prediction problem. Multi-task learning allows correlation between geographical regions. 

The region from which datasets are sourced often impacts performance of systems. For example, \citet{chapman2005classifying} train their system on chief complaints from Pennsylvania, and test it on those from Utah. \citet{ofoghi2016towards} use a negative dataset from Australia while a positive dataset is from London.
 \citet{sadilek2012predicting} predict individual infections based on: indicators within text and geo-tagging in terms of locations and co-locations with people and infected people. Therefore, they use a CRF to make their estimations. In addition to textual features, each observed instance in the CRF is characterised by: (a) number of collocations in past 7 days with anyone, (b) number of collocations in past seven days with people who have reported illnesses, and (c) number of collocations in past seven days with friends who have reported illnesses.  \citet{gomide2011dengue} estimate the counts of dengue incidences in different geographical areas for a dataset of tweets from Brazil. Then, depending on these counts, they use a  spatial clustering algorithm that creates clusters of cities depending on their physical proximity and number of dengue incidences. This approach is helpful to understand the spread of the disease.
On the other hand, \cite{shao2017efficient} use co-mentions in tweets to create a social network of users. They then apply scan statistics to identify health events in the network. In this case, the notion of space is a virtual network on social media.

\section{Evaluation}
\label{sec:eval}
This section presents evaluation methodologies used to validate the performance of epidemic intelligence. Evaluation using manually labeled datasets is common, as is the case for most supervised classification tasks. In addition, the following trends emerge in past work in terms of evaluation methodologies:
\begin{enumerate}
    \item \textbf{Correlation with publicly available health data}: Health data may be publicly available in terms of counts of infections or known health outbreaks. Either of the two can be used to evaluate epidemic intelligence. In general, the framework to evaluate epidemic intelligence comprises following steps~\cite{lampos2017enhancing,lamb2013separating}:
    \begin{enumerate}
    \item To evaluate health mention classification, they report classification performance on a labeled dataset, annotated using manual, automatic or combined annotation strategies.
    \item This may be followed by correlation with publicly available counts for infections. This is often done by training appropriate regression models which predict infection counts. Alternatively, the outbreaks returned by health event detection could be compared against health events that have been known to happen from other sources such as news.
    \end{enumerate}
\citet{chen2016syndromic} use Pan American Health Organization (PAHO) case counts to validate their predictions. \citet{aamer2016syndromic} report the number of incidents and top terms discovered by three sliding window configurations. \citet{velardi2014twitter} identify a set of terms of health-related words and download tweets containing these words. These counts are then correlated with ILI counts.  \citet{huang2016syndromic} experiment with Lyme and Flu, and show the correlation for two series of CDC counts: counts of the current week and counts of the next week. They show that outbreaks can be predicted a week in advance using tweet streams.  \citet{aramaki2011twitter} first evaluate the classification performance on 10-fold cross-validation. They then correlate the Google Flu trends with the disease outbreaks as predicted for the test datasets. They observe that excessive news reporting may lead to false alerts from social media. Similarly, \citet{pervaiz2012flubreaks} present a comparison of three classes of algorithms for epidemic detection from Google Flu Trends: normal distribution algorithms, Poisson distribution algorithms and negative binomial distribution algorithms.
\item \textbf{Validation on multiple datasets}:~\citet{ofoghi2016towards} experiment with two datasets: a positive dataset which contains a syndromic outbreak and a negative dataset which is from a year before the outbreak. They then validate if the outbreak in the positive dataset gets detected, as well as no outbreak is detected in the negative dataset. Similarly, to validate their approach of distorting and partitioning word embeddings, \citet{karisani2018did} report their results on multiple illnesses. \citet{hayate2016forecasting} create three datasets for three seasons/spells of influenza and show correlation with the reported flu counts. They train the model on one season while testing on another. \citet{yates2014framework} report their observations for two medical conditions: allergies and flu. 
\item \textbf{Evaluation of components}: Often, epidemic intelligence may consist of components. This is typical in the case of pipeline-based approaches. Therefore, \citet{yepes2015investigating}, a pipeline-based approach, report performance of every stage of the pipeline, namely, (i) Performance of medical NER and geotagger on a manually labeled dataset, (ii) Top medical terms in tweets that are extracted, (iii) seven day rolling average for three cities, namely, New York, London and Chicago. Components may also be added to a downstream task. For example, \citet{kanouchi2015caught} present an approach to identify target of a personal health mention. To validate that this is useful, they interface it with a downstream task of health event detection (referred to as `episode prediction' in their paper). They show that performing target identification before a personal health mention detection results in an improved performance of health event detection.
\item \textbf{Qualitative evaluation}: \citet{ginsberg2009detecting} reports top topics in health-related search queries across different localities. Similarly, ~\citet{paul2011you} describe topics generated by the topic model inx terms of word clusters.
\end{enumerate}

\section{Open Research Avenues}
\label{sec:openres}
Based on this survey, we now identify the following research directions for epidemic intelligence using computational linguistic techniques. These avenues represent enhancements in three directions: (a) Quality of the surveillance output (by mitigating false alerts), (b) Timeliness of the surveillance output (to achieve near-real-time indicators), and (c) Coverage of the system (in terms of factoring relationships between symptoms/syndromes).
\begin{enumerate}
\item \textbf{Mitigation of False Alerts}: \citet{ginsberg2009detecting} observe possible false alerts due to reliance on social media. This implies that improved precision without impacting recall is a useful research avenue for epidemic intelligence. Towards this, we note two possibilities. The first possibility is in terms of detection of spam. Due to possibly malicious intent of social media users, false information may be published in social media posts. The second possibility is in terms of figurative language. Since many symptoms have figurative usages (`\textit{my naughty kids almost gave me a heart-attack}'), separating figurative language from literal language may prove to be beneficial. Additional checks such as these could be incorporated into existing epidemic intelligence approaches, to avoid false alerts in general.
\item \textbf{Opportunities for Near-Real-time Indicators}: Epidemic intelligence can be useful to manage ambulance networks in times of an outbreak~\cite{sparks2010understanding}. It would be useful to investigate if social media text provides real-time signals to identify outbreaks at sub-daily intervals. \citet{velasco2014social} enlist challenges in integration of event-based techniques for social media surveillance.
\item \textbf{Overlapping Syndromes and Symptoms}: Past work considers different illness/syndromes as merely different datasets or systems against which experiments are to be run. However, many of these syndromes may be related to each other. An interesting future direction is to consider how syndromes are similar to one another in terms of their symptoms. It follows that epidemic intelligence for physical illnesses may be combined with mental health surveillance or animal health surveillance. The former is crucial since mental health conditions may involve physical symptoms, the latter assumes importance due to zoonotic diseases that may be transferred from animals to humans. An initial work in the direction of animal monitoring is by \citet{welvaert2017limits} who use social media as a monitoring tool for exotic species. 
\end{enumerate}

\section{Conclusions}
\label{sec:concl}
Text-based epidemic intelligence has received attention due to the information and timeliness of textual data on the web. Techniques involving different levels of sophistication of computational linguistic approaches have been reported. In this paper, we survey these past approaches. We first introduce textual datasets, highlighting the strengths and challenges in each. We note that ontologies that capture medical concepts have been valuable for text-based epidemic intelligence. Also, since social media is an accessible medium today, social media text such as tweets also provide an opportunity for text-based epidemic intelligence. 

We then view past work in terms of health mention classification (which deals with detecting syndromes in individual textual units) and health event detection (which deals with detecting outbreaks using a collection of textual units). Health mention classification techniques may use ontologies, pipelines of NLP components, statistical classifiers with task-specific features or neural network architectures. Advances in natural language processing and machine learning have been employed for newer approaches of health mention classification. In terms of health event detection, we investigate how large volumes of text have been used to detect health events relevant to a community, and how geographical information has been used to fine-tune these predictions. Based on our survey, we believe that avenues for future work in this area lie in terms of improving the quality (by mitigating false alerts), the efficiency (by making health event detection as real-time as possible) and the coverage of epidemic intelligence (by combining related symptoms).

We hope that our computational linguistic perspective to epidemic intelligence serves as a useful resource for computational linguists and health practitioners alike.
\bibliographystyle{acl_natbib}
\bibliography{our-bibliography.bib}

\begin{thebibliography}{73}
\expandafter\ifx\csname natexlab\endcsname\relax\def\natexlab#1{#1}\fi

\bibitem[{Aamer et~al.(2016)Aamer, Ofoghi, and Verspoor}]{aamer2016syndromic}
Hafsah Aamer, Bahadorreza Ofoghi, and Karin Verspoor. 2016.
\newblock Syndromic surveillance through measuring lexical shift in emergency
  department chief complaint texts.
\newblock In \emph{Proceedings of the Australasian Language Technology
  Association Workshop 2016}, pages 45--53.

\bibitem[{Adam et~al.(2017)Adam, Jonnagaddala, Han-Chen, Batongbacal, Almeida,
  Zhu, Yang, Mundekkat, Badman, Chughtai et~al.}]{adam2017zikahack}
Dillon~C Adam, Jitendra Jonnagaddala, Daniel Han-Chen, Sean Batongbacal, Luan
  Almeida, Jing~Z Zhu, Jenny~J Yang, Jumail~M Mundekkat, Steven Badman, Abrar
  Chughtai, et~al. 2017.
\newblock Zikahack 2016: A digital disease detection competition.
\newblock In \emph{Proceedings of the International Workshop on Digital Disease
  Detection using Social Media 2017 (DDDSM-2017)}, pages 39--46.

\bibitem[{Al-garadi et~al.(2016)Al-garadi, Khan, Varathan, Mujtaba, and
  Al-Kabsi}]{al2016using}
Mohammed~Ali Al-garadi, Muhammad~Sadiq Khan, Kasturi~Dewi Varathan, Ghulam
  Mujtaba, and Abdelkodose~M Al-Kabsi. 2016.
\newblock Using online social networks to track a pandemic: A systematic
  review.
\newblock \emph{Journal of biomedical informatics}, 62:1--11.

\bibitem[{Alicino et~al.(2015)Alicino, Bragazzi, Faccio, Amicizia, Panatto,
  Gasparini, Icardi, and Orsi}]{alicino2015assessing}
Cristiano Alicino, Nicola~Luigi Bragazzi, Valeria Faccio, Daniela Amicizia,
  Donatella Panatto, Roberto Gasparini, Giancarlo Icardi, and Andrea Orsi.
  2015.
\newblock Assessing ebola-related web search behaviour: insights and
  implications from an analytical study of google trends-based query volumes.
\newblock \emph{Infectious diseases of poverty}, 4(1):54.

\bibitem[{Aramaki et~al.(2011)Aramaki, Maskawa, and
  Morita}]{aramaki2011twitter}
Eiji Aramaki, Sachiko Maskawa, and Mizuki Morita. 2011.
\newblock Twitter catches the flu: detecting influenza epidemics using twitter.
\newblock In \emph{Proceedings of the conference on empirical methods in
  natural language processing}, pages 1568--1576. Association for Computational
  Linguistics.

\bibitem[{Benton et~al.(2017)Benton, Coppersmith, and
  Dredze}]{benton2017ethical}
Adrian Benton, Glen Coppersmith, and Mark Dredze. 2017.
\newblock Ethical research protocols for social media health research.
\newblock In \emph{Proceedings of the First ACL Workshop on Ethics in Natural
  Language Processing}, pages 94--102.

\bibitem[{Bernardo et~al.(2013)Bernardo, Rajic, Young, Robiadek, Pham, and
  Funk}]{bernardo2013scoping}
Theresa~Marie Bernardo, Andrijana Rajic, Ian Young, Katie Robiadek, Mai~T Pham,
  and Julie~A Funk. 2013.
\newblock Scoping review on search queries and social media for disease
  surveillance: a chronology of innovation.
\newblock \emph{Journal of medical Internet research}, 15(7).

\bibitem[{Bertaud-Gounot et~al.(2012)Bertaud-Gounot, Duvauferrier, and
  Burgun}]{bertaud2012ontology}
Val{\'e}rie Bertaud-Gounot, R{\'e}gis Duvauferrier, and Anita Burgun. 2012.
\newblock Ontology and medical diagnosis.
\newblock \emph{Informatics for Health and Social Care}, 37(2):51--61.

\bibitem[{Blei et~al.(2003)Blei, Ng, and Jordan}]{bleispaper}
David~M Blei, Andrew~Y Ng, and Michael~I Jordan. 2003.
\newblock Latent dirichlet allocation.
\newblock \emph{Journal of machine Learning research}, 3(Jan):993--1022.

\bibitem[{Bodenreider(2004)}]{bodenreider2004unified}
Olivier Bodenreider. 2004.
\newblock The unified medical language system (umls): integrating biomedical
  terminology.
\newblock \emph{Nucleic acids research}, 32(suppl\_1):D267--D270.

\bibitem[{Boyle et~al.(2011)Boyle, Sparks, Keijzers, Crilly, Lind, and
  Ryan}]{boyle2011prediction}
Justin~R Boyle, Ross~S Sparks, Gerben~B Keijzers, Julia~L Crilly, James~F Lind,
  and Louise~M Ryan. 2011.
\newblock Prediction and surveillance of influenza epidemics.
\newblock \emph{Medical Journal of Australia}, 194(4):S28.

\bibitem[{Brownstein et~al.(2009)Brownstein, Freifeld, and
  Madoff}]{brownstein2009digital}
John~S Brownstein, Clark~C Freifeld, and Lawrence~C Madoff. 2009.
\newblock Digital disease detection—harnessing the web for public health
  surveillance.
\newblock \emph{New England Journal of Medicine}, 360(21):2153--2157.

\bibitem[{Chapman et~al.(2005)Chapman, Christensen, Wagner, Haug, Ivanov,
  Dowling, and Olszewski}]{chapman2005classifying}
Wendy~W Chapman, Lee~M Christensen, Michael~M Wagner, Peter~J Haug, Oleg
  Ivanov, John~N Dowling, and Robert~T Olszewski. 2005.
\newblock Classifying free-text triage chief complaints into syndromic
  categories with natural language processing.
\newblock \emph{Artificial Intelligence in Medicine}, 33(1):31--40.

\bibitem[{Charles-Smith et~al.(2015)Charles-Smith, Reynolds, Cameron, Conway,
  Lau, Olsen, Pavlin, Shigematsu, Streichert, Suda et~al.}]{charles2015using}
Lauren~E Charles-Smith, Tera~L Reynolds, Mark~A Cameron, Mike Conway, Eric~HY
  Lau, Jennifer~M Olsen, Julie~A Pavlin, Mika Shigematsu, Laura~C Streichert,
  Katie~J Suda, et~al. 2015.
\newblock Using social media for actionable disease surveillance and outbreak
  management: a systematic literature review.
\newblock \emph{PloS one}, 10(10):e0139701.

\bibitem[{Chen et~al.(2016)Chen, Hossain, Butler, Ramakrishnan, and
  Prakash}]{chen2016syndromic}
Liangzhe Chen, KSM~Tozammel Hossain, Patrick Butler, Naren Ramakrishnan, and
  B~Aditya Prakash. 2016.
\newblock Syndromic surveillance of flu on twitter using weakly supervised
  temporal topic models.
\newblock \emph{Data Mining and Knowledge Discovery}, 30(3):681--710.

\bibitem[{Collier et~al.(2007)Collier, Kawazoe, Jin, Shigematsu, Dien
  et~al.}]{collier2007biocaster}
N~Collier, A~Kawazoe, L~Jin, M~Shigematsu, D~Dien, et~al. 2007.
\newblock The biocaster ontology: A multilingual ontology for infectious
  disease outbreak surveillance: Rationale, design and challenges.
\newblock \emph{Journal of Language Resources and Evaluation}, pages 405--413.

\bibitem[{Collier et~al.(2010)Collier, Goodwin, McCrae, Doan, Kawazoe, Conway,
  Kawtrakul, Takeuchi, and Dien}]{collier2010ontology}
Nigel Collier, Reiko~Matsuda Goodwin, John McCrae, Son Doan, Ai~Kawazoe, Mike
  Conway, Asanee Kawtrakul, Koichi Takeuchi, and Dinh Dien. 2010.
\newblock An ontology-driven system for detecting global health events.
\newblock In \emph{Proceedings of the 23rd International Conference on
  Computational Linguistics}, pages 215--222. Association for Computational
  Linguistics.

\bibitem[{Conway et~al.(2011)Conway, Dowling, and
  Chapman}]{conway2011developing}
Mike Conway, John Dowling, and Wendy Chapman. 2011.
\newblock Developing an application ontology for mining free text clinical
  reports: the extended syndromic surveillance ontology.
\newblock In \emph{3rd International Workshop on Health Document Text Mining
  and Information Analysis (LOUHI 2011)}, pages 75--82. Citeseer.

\bibitem[{Conway et~al.(2013)Conway, Dowling, and Chapman}]{conway2013using}
Mike Conway, John~N Dowling, and Wendy~W Chapman. 2013.
\newblock Using chief complaints for syndromic surveillance: a review of chief
  complaint based classifiers in north america.
\newblock \emph{Journal of Biomedical Informatics}, 46(4):734--743.

\bibitem[{Crub{\'e}zy et~al.(2005)Crub{\'e}zy, O'Connor, Pincus, Musen, and
  Buckeridge}]{crubezy2005ontology}
Monica Crub{\'e}zy, Martin O'Connor, Zachary Pincus, Mark~A Musen, and David~L
  Buckeridge. 2005.
\newblock Ontology-centered syndromic surveillance for bioterrorism.
\newblock \emph{IEEE Intelligent Systems}, 20(5):26--35.

\bibitem[{Dai et~al.(2017)Dai, Bikdash, and Meyer}]{dai2017social}
Xiangfeng Dai, Marwan Bikdash, and Bradley Meyer. 2017.
\newblock From social media to public health surveillance: Word embedding based
  clustering method for twitter classification.
\newblock In \emph{SoutheastCon, 2017}, pages 1--7. IEEE.

\bibitem[{Doan et~al.(2008)Doan, Kawazoe, Collier et~al.}]{doan2008global}
Son Doan, Ai~Kawazoe, Nigel Collier, et~al. 2008.
\newblock Global health monitor-a web-based system for detecting and mapping
  infectious diseases.
\newblock In \emph{Proceedings of the Third International Joint Conference on
  Natural Language Processing}.

\bibitem[{Freifeld et~al.(2008)Freifeld, Mandl, Reis, and
  Brownstein}]{freifeld2008healthmap}
Clark~C Freifeld, Kenneth~D Mandl, Ben~Y Reis, and John~S Brownstein. 2008.
\newblock Healthmap: global infectious disease monitoring through automated
  classification and visualization of internet media reports.
\newblock \emph{Journal of the American Medical Informatics Association},
  15(2):150--157.

\bibitem[{Fung et~al.(2015)Fung, Tse, and Fu}]{fung2015use}
Isaac Chun-Hai Fung, Zion Tsz~Ho Tse, and King-Wa Fu. 2015.
\newblock The use of social media in public health surveillance.
\newblock \emph{Western Pacific surveillance and response journal: WPSAR},
  6(2):3.

\bibitem[{Ginsberg et~al.(2009)Ginsberg, Mohebbi, Patel, Brammer, Smolinski,
  and Brilliant}]{ginsberg2009detecting}
Jeremy Ginsberg, Matthew~H Mohebbi, Rajan~S Patel, Lynnette Brammer, Mark~S
  Smolinski, and Larry Brilliant. 2009.
\newblock Detecting influenza epidemics using search engine query data.
\newblock \emph{Nature}, 457(7232):1012.

\bibitem[{Gomide et~al.(2011)Gomide, Veloso, Meira~Jr, Almeida, Benevenuto,
  Ferraz, and Teixeira}]{gomide2011dengue}
Jana{\'\i}na Gomide, Adriano Veloso, Wagner Meira~Jr, Virg{\'\i}lio Almeida,
  Fabr{\'\i}cio Benevenuto, Fernanda Ferraz, and Mauro Teixeira. 2011.
\newblock Dengue surveillance based on a computational model of spatio-temporal
  locality of twitter.
\newblock In \emph{Proceedings of the 3rd International Web Science
  Conference}, page~3. ACM.

\bibitem[{Gruber(1993)}]{ontologydef}
Thomas~R Gruber. 1993.
\newblock A translation approach to portable ontology specifications.
\newblock \emph{Knowledge acquisition}, 5(2):199--220.

\bibitem[{Hayate et~al.(2016)Hayate, Wakamiya, and
  Aramaki}]{hayate2016forecasting}
ISO Hayate, Shoko Wakamiya, and Eiji Aramaki. 2016.
\newblock Forecasting word model: Twitter-based influenza surveillance and
  prediction.
\newblock In \emph{Proceedings of the 26th International Conference on
  Computational Linguistics}, pages 76--86.

\bibitem[{Henning(2004)}]{henning2004syndromic}
Kelly~J Henning. 2004.
\newblock What is syndromic surveillance?
\newblock \emph{Morbidity and Mortality Weekly Report}, pages 7--11.

\bibitem[{Hopkins et~al.(2017)Hopkins, Tong, Burkom, Akkina, Berezowski,
  Shigematsu, Finley, Painter, Gamache, Vilas et~al.}]{hopkins2017practitioner}
Richard~S Hopkins, Catherine~C Tong, Howard~S Burkom, Judy~E Akkina, John
  Berezowski, Mika Shigematsu, Patrick~D Finley, Ian Painter, Roland Gamache,
  Victor J Del~Rio Vilas, et~al. 2017.
\newblock A practitioner-driven research agenda for syndromic surveillance.
\newblock \emph{Public Health Reports}, 132(1\_suppl):116S--126S.

\bibitem[{Huang et~al.(2016)Huang, MacKinlay, and Yepes}]{huang2016syndromic}
Pin Huang, Andrew MacKinlay, and Antonio~Jimeno Yepes. 2016.
\newblock Syndromic surveillance using generic medical entities on twitter.
\newblock In \emph{Proceedings of the Australasian Language Technology
  Association Workshop 2016}, pages 35--44.

\bibitem[{Hulth et~al.(2009)Hulth, Rydevik, and Linde}]{hulth2009web}
Anette Hulth, Gustaf Rydevik, and Annika Linde. 2009.
\newblock Web queries as a source for syndromic surveillance.
\newblock \emph{PloS one}, 4(2):e4378.

\bibitem[{Jiang et~al.(2016)Jiang, Calix, and Gupta}]{jiang2016construction}
Keyuan Jiang, Ricardo Calix, and Matrika Gupta. 2016.
\newblock Construction of a personal experience tweet corpus for health
  surveillance.
\newblock In \emph{Proceedings of the 15th Workshop on Biomedical Natural
  Language Processing}, pages 128--135.

\bibitem[{Joshi et~al.(2017)Joshi, Bhattacharyya, and
  Ahire}]{joshi2017sentiment}
Aditya Joshi, Pushpak Bhattacharyya, and Sagar Ahire. 2017.
\newblock Sentiment resources: Lexicons and datasets.
\newblock In \emph{A Practical Guide to Sentiment Analysis}, pages 85--106.
  Springer.

\bibitem[{Kanouchi et~al.(2015)Kanouchi, Komachi, Okazaki, Aramaki, and
  Ishikawa}]{kanouchi2015caught}
Shin Kanouchi, Mamoru Komachi, Naoaki Okazaki, Eiji Aramaki, and Hiroshi
  Ishikawa. 2015.
\newblock Who caught a cold?-identifying the subject of a symptom.
\newblock In \emph{Proceedings of the 53rd Annual Meeting of the Association
  for Computational Linguistics and the 7th International Joint Conference on
  Natural Language Processing (Volume 1: Long Papers)}, volume~1, pages
  1660--1670.

\bibitem[{Karimi et~al.(2015)Karimi, Wang, Metke-Jimenez, Gaire, and
  Paris}]{karimi2015text}
Sarvnaz Karimi, Chen Wang, Alejandro Metke-Jimenez, Raj Gaire, and Cecile
  Paris. 2015.
\newblock Text and data mining techniques in adverse drug reaction detection.
\newblock \emph{ACM Computing Surveys}, 47(4):56.

\bibitem[{Karisani and Agichtein(2018)}]{karisani2018did}
Payam Karisani and Eugene Agichtein. 2018.
\newblock Did you really just have a heart attack?: Towards robust detection of
  personal health mentions in social media.
\newblock In \emph{Proceedings of the 2018 World Wide Web Conference on World
  Wide Web}, pages 137--146. International World Wide Web Conferences Steering
  Committee.

\bibitem[{Lamb et~al.(2013)Lamb, Paul, and Dredze}]{lamb2013separating}
Alex Lamb, Michael~J Paul, and Mark Dredze. 2013.
\newblock Separating fact from fear: Tracking flu infections on twitter.
\newblock In \emph{Proceedings of the 2013 Conference of the North American
  Chapter of the Association for Computational Linguistics: Human Language
  Technologies}, pages 789--795.

\bibitem[{Lampos et~al.(2017)Lampos, Zou, and Cox}]{lampos2017enhancing}
Vasileios Lampos, Bin Zou, and Ingemar~Johansson Cox. 2017.
\newblock Enhancing feature selection using word embeddings: The case of flu
  surveillance.
\newblock In \emph{Proceedings of the 26th International Conference on World
  Wide Web}, pages 695--704. International World Wide Web Conferences Steering
  Committee.

\bibitem[{Larsen et~al.(2015)Larsen, Boonstra, Batterham, O’Dea, Paris, and
  Christensen}]{larsen2015we}
Mark~E Larsen, Tjeerd~W Boonstra, Philip~J Batterham, Bridianne O’Dea, Cecile
  Paris, and Helen Christensen. 2015.
\newblock We feel: mapping emotion on twitter.
\newblock \emph{IEEE journal of Biomedical and Health Informatics},
  19(4):1246--1252.

\bibitem[{Lejeune et~al.(2010)Lejeune, Doucet, Yangarber, and
  Lucas}]{lejeune2010filtering}
Ga{\"e}l Lejeune, Antoine Doucet, Roman Yangarber, and Nadine Lucas. 2010.
\newblock Filtering news for epidemic surveillance: towards processing more
  languages with fewer resources.
\newblock In \emph{4th International workshop on cross-lingual information
  access}, pages 8--pages.

\bibitem[{Lindberg et~al.(1993)Lindberg, Humphreys, and
  McCray}]{lindberg1993unified}
Donald~AB Lindberg, Betsy~L Humphreys, and Alexa~T McCray. 1993.
\newblock The unified medical language system.
\newblock \emph{Methods of information in medicine}, 32(04):281--291.

\bibitem[{Lu et~al.(2009)Lu, Chen, Zeng, King, Shih, Wu, and
  Hsiao}]{lu2009multilingual}
Hsin-Min Lu, Hsinchun Chen, Daniel Zeng, Chwan-Chuen King, Fuh-Yuan Shih,
  Tsung-Shu Wu, and Jin-Yi Hsiao. 2009.
\newblock Multilingual chief complaint classification for syndromic
  surveillance: an experiment with chinese chief complaints.
\newblock \emph{International Journal of Medical Informatics}, 78(5):308--320.

\bibitem[{Manning et~al.(1999)Manning, Manning, and
  Sch{\"u}tze}]{manning1999foundations}
Christopher~D Manning, Christopher~D Manning, and Hinrich Sch{\"u}tze. 1999.
\newblock \emph{Foundations of statistical natural language processing}.
\newblock MIT press.

\bibitem[{N{\'e}v{\'e}ol et~al.(2009)N{\'e}v{\'e}ol, Kim, Wilbur, and
  Lu}]{neveol2009exploring}
Aur{\'e}lie N{\'e}v{\'e}ol, Won Kim, W~John Wilbur, and Zhiyong Lu. 2009.
\newblock Exploring two biomedical text genres for disease recognition.
\newblock In \emph{Proceedings of the Workshop on current trends in Biomedical
  Natural Language Processing}, pages 144--152. Association for Computational
  Linguistics.

\bibitem[{Ofoghi et~al.(2016)Ofoghi, Mann, and Verspoor}]{ofoghi2016towards}
Bahadorreza Ofoghi, Meghan Mann, and Karin Verspoor. 2016.
\newblock Towards early discovery of salient health threats: A social media
  emotion classification technique.
\newblock In \emph{Biocomputing 2016: Proceedings of the Pacific Symposium},
  pages 504--515. World Scientific.

\bibitem[{Okhmatovskaia et~al.(2009)Okhmatovskaia, Chapman, Collier, Espino,
  and Buckeridge}]{okhmatovskaia2009sso}
A~Okhmatovskaia, W~Chapman, N~Collier, J~Espino, and DL~Buckeridge. 2009.
\newblock Sso: the syndromic surveillance ontology.
\newblock In \emph{Proceedings of the International Society for Disease
  Surveillance}.

\bibitem[{Olszewski(2003)}]{olszewski2003bayesian}
Robert~T Olszewski. 2003.
\newblock Bayesian classification of triage diagnoses for the early detection
  of epidemics.
\newblock In \emph{International Florida Artificial Intelligence Research
  Society Conference}, pages 412--416.

\bibitem[{Paul and Dredze(2011)}]{paul2011you}
Michael~J Paul and Mark Dredze. 2011.
\newblock You are what you tweet: Analyzing twitter for public health.
\newblock In \emph{International AAAI Conference on Web and Social Media},
  volume~20, pages 265--272.

\bibitem[{Paul and Dredze(2012)}]{paul2012model}
Michael~J Paul and Mark Dredze. 2012.
\newblock A model for mining public health topics from twitter.
\newblock \emph{Health}, 11:16--6.

\bibitem[{Pervaiz et~al.(2012)Pervaiz, Pervaiz, Rehman, and
  Saif}]{pervaiz2012flubreaks}
Fahad Pervaiz, Mansoor Pervaiz, Nabeel~Abdur Rehman, and Umar Saif. 2012.
\newblock Flubreaks: early epidemic detection from google flu trends.
\newblock \emph{Journal of medical Internet research}, 14(5).

\bibitem[{Rosewell et~al.(2013)Rosewell, Ropa, Randall, Dagina, Hurim, Bieb,
  Datta, Ramamurthy, Mola, Zwi et~al.}]{rosewell2013mobile}
Alexander Rosewell, Berry Ropa, Heather Randall, Rosheila Dagina, Samuel Hurim,
  Sibauk Bieb, Siddhartha Datta, Sundar Ramamurthy, Glen Mola, Anthony~B Zwi,
  et~al. 2013.
\newblock Mobile phone--based syndromic surveillance system, papua new guinea.
\newblock \emph{Emerging Infectious Diseases}, 19(11):1811.

\bibitem[{Sadilek et~al.(2012)Sadilek, Kautz, and
  Silenzio}]{sadilek2012predicting}
Adam Sadilek, Henry~A Kautz, and Vincent Silenzio. 2012.
\newblock Predicting disease transmission from geo-tagged micro-blog data.
\newblock In \emph{Conference on Artificial Intelligence (AAAI)}, pages
  136--142.

\bibitem[{Sarker et~al.(2015)Sarker, Ginn, Nikfarjam, O’Connor, Smith,
  Jayaraman, Upadhaya, and Gonzalez}]{sarker2015utilizing}
Abeed Sarker, Rachel Ginn, Azadeh Nikfarjam, Karen O’Connor, Karen Smith,
  Swetha Jayaraman, Tejaswi Upadhaya, and Graciela Gonzalez. 2015.
\newblock Utilizing social media data for pharmacovigilance: a review.
\newblock \emph{Journal of biomedical informatics}, 54:202--212.

\bibitem[{Sarker et~al.(2016)Sarker, Nikfarjam, and
  Gonzalez}]{sarker2016social}
Abeed Sarker, Azadeh Nikfarjam, and Graciela Gonzalez. 2016.
\newblock Social media mining shared task workshop.
\newblock In \emph{Biocomputing 2016: Proceedings of the Pacific Symposium},
  pages 581--592. World Scientific.

\bibitem[{Shao et~al.(2017)Shao, Li, Chen, Huang, Zhang, and
  Chen}]{shao2017efficient}
Minglai Shao, Jianxin Li, Feng Chen, Hongyi Huang, Shuai Zhang, and Xunxun
  Chen. 2017.
\newblock An efficient approach to event detection and forecasting in dynamic
  multivariate social media networks.
\newblock In \emph{Proceedings of the 26th International Conference on World
  Wide Web}, pages 1631--1639. International World Wide Web Conferences
  Steering Committee.

\bibitem[{Sparks et~al.(2010{\natexlab{a}})Sparks, Carter, Graham, Muscatello,
  Churches, Kaldor, Turner, Zheng, and Ryan}]{sparks2010understanding}
Ross Sparks, Chris Carter, Petra Graham, David Muscatello, Tim Churches, Jill
  Kaldor, Robyn Turner, Wei Zheng, and Louise Ryan. 2010{\natexlab{a}}.
\newblock Understanding sources of variation in syndromic surveillance for
  early warning of natural or intentional disease outbreaks.
\newblock \emph{IIE Transactions}, 42(9):613--631.

\bibitem[{Sparks et~al.(2010{\natexlab{b}})Sparks, Keighley, and
  Muscatello}]{sparks2010exponentially}
Ross Sparks, Tim Keighley, and David Muscatello. 2010{\natexlab{b}}.
\newblock Exponentially weighted moving average plans for detecting unusual
  negative binomial counts.
\newblock \emph{IIE Transactions}, 42(10):721--733.

\bibitem[{Sparks et~al.(2017)Sparks, Robinson, Power, Cameron, and
  Woolford}]{sparks2017investigation}
Ross~S Sparks, Bella Robinson, Robert Power, Mark Cameron, and Sam Woolford.
  2017.
\newblock An investigation into social media syndromic monitoring.
\newblock \emph{Communications in Statistics-Simulation and Computation},
  46(8):5901--5923.

\bibitem[{Velardi et~al.(2014)Velardi, Stilo, Tozzi, and
  Gesualdo}]{velardi2014twitter}
Paola Velardi, Giovanni Stilo, Alberto~E Tozzi, and Francesco Gesualdo. 2014.
\newblock Twitter mining for fine-grained syndromic surveillance.
\newblock \emph{Artificial Intelligence in Medicine}, 61(3):153--163.

\bibitem[{Velasco et~al.(2014)Velasco, Agheneza, Denecke, Kirchner, and
  Eckmanns}]{velasco2014social}
Edward Velasco, Tumacha Agheneza, Kerstin Denecke, Goeran Kirchner, and Tim
  Eckmanns. 2014.
\newblock Social media and internet-based data in global systems for public
  health surveillance: A systematic review.
\newblock \emph{The Milbank Quarterly}, 92(1):7--33.

\bibitem[{Wagner et~al.(2011)Wagner, Moore, and Aryel}]{wagner2011handbook}
Michael~M Wagner, Andrew~W Moore, and Ron~M Aryel. 2011.
\newblock \emph{Handbook of Biosurveillance}.
\newblock Elsevier.

\bibitem[{Wang et~al.(2017)Wang, Singh, Tang, and Dai}]{wang2017using}
Chen-Kai Wang, Onkar Singh, Zhao-Li Tang, and Hong-Jie Dai. 2017.
\newblock Using a recurrent neural network model for classification of tweets
  conveyed influenza-related information.
\newblock In \emph{Proceedings of the International Workshop on Digital Disease
  Detection using Social Media 2017 (DDDSM-2017)}, pages 33--38.

\bibitem[{Wang et~al.(2014)Wang, Paul, and Dredze}]{wang2014exploring}
Shiliang Wang, Michael~J Paul, and Mark Dredze. 2014.
\newblock Exploring health topics in chinese social media: An analysis of sina
  weibo.
\newblock In \emph{AAAI Workshop on the World Wide Web and Public Health
  Intelligence}, volume~31, page~59.

\bibitem[{Weissenbacher et~al.(2018)Weissenbacher, Sarker, Paul, and
  Gonzalez-Hernandez}]{weissenbacher2018overview}
Davy Weissenbacher, Abeed Sarker, Michael~J Paul, and Graciela
  Gonzalez-Hernandez. 2018.
\newblock Overview of the third social media mining for health (smm4h) shared
  tasks at emnlp 2018.
\newblock In \emph{Proceedings of the 2018 EMNLP Workshop SMM4H: The 3rd Social
  Media Mining for Health Applications Workshop and Shared Task}, pages 13--16.

\bibitem[{Welvaert et~al.(2017)Welvaert, Al-Ghattas, Cameron, and
  Caley}]{welvaert2017limits}
Marijke Welvaert, Omar Al-Ghattas, Mark Cameron, and Peter Caley. 2017.
\newblock Limits of use of social media for monitoring biosecurity events.
\newblock \emph{PloS one}, 12(2):e0172457.

\bibitem[{Woo et~al.(2016)Woo, Cho, Shim, Lee, Lee, and
  Kim}]{woo2016estimating}
Hyekyung Woo, Youngtae Cho, Eunyoung Shim, Jong-Koo Lee, Chang-Gun Lee, and
  Seong~Hwan Kim. 2016.
\newblock Estimating influenza outbreaks using both search engine query data
  and social media data in south korea.
\newblock \emph{Journal of medical Internet research}, 18(7).

\bibitem[{Yan et~al.(2006)Yan, Zeng, and Chen}]{yan2006review}
Ping Yan, Daniel Zeng, and Hsinchun Chen. 2006.
\newblock A review of public health syndromic surveillance systems.
\newblock In \emph{International Conference on Intelligence and Security
  Informatics}, pages 249--260. Springer.

\bibitem[{Yan et~al.(2017)Yan, Chughtai, and Macintyre}]{yan2017utility}
SJ~Yan, AA~Chughtai, and CR~Macintyre. 2017.
\newblock Utility and potential of rapid epidemic intelligence from
  internet-based sources.
\newblock \emph{International Journal of Infectious Diseases}, 63:77--87.

\bibitem[{Yangarber et~al.(2008)Yangarber, Von~Etter, and
  Steinberger}]{yangarber2008content}
Roman Yangarber, Peter Von~Etter, and Ralf Steinberger. 2008.
\newblock Content collection and analysis in the domain of epidemiology.
\newblock In \emph{Proceedings of DrMED-2008: International Workshop on
  Describing Medical Web Resources}.

\bibitem[{Yates et~al.(2014)Yates, Parker, Goharian, and
  Frieder}]{yates2014framework}
Andrew Yates, Jon Parker, Nazli Goharian, and Ophir Frieder. 2014.
\newblock A framework for public health surveillance.
\newblock In \emph{Language Resources and Evaluation Conference}, pages
  475--482.

\bibitem[{Yepes et~al.(2015)Yepes, MacKinlay, and Han}]{yepes2015investigating}
Antonio~Jimeno Yepes, Andrew MacKinlay, and Bo~Han. 2015.
\newblock Investigating public health surveillance using twitter.
\newblock \emph{Proceedings of BioNLP}, pages 164--170.

\bibitem[{Zou et~al.(2018)Zou, Lampos, and Cox}]{zou2018multi}
Bin Zou, Vasileios Lampos, and Ingemar Cox. 2018.
\newblock Multi-task learning improves disease models from web search.
\newblock In \emph{Proceedings of the 2018 World Wide Web Conference}, pages
  87--96. International World Wide Web Conferences Steering Committee.

\end{thebibliography}
\end{document}